\documentclass{article}

\usepackage[preprint]{neurips_2024}

\usepackage[utf8]{inputenc} 
\usepackage[T1]{fontenc}    
\usepackage{url}            
\usepackage{booktabs}       
\usepackage{amsfonts}       
\usepackage{nicefrac}       
\usepackage{microtype}      
\usepackage{xcolor}         

\usepackage{amsmath}
\usepackage{amssymb}
\usepackage{mathtools}
\usepackage{amsthm}

\usepackage{multirow}
\usepackage{comment} 
\usepackage{enumitem}

\usepackage{times}
\usepackage{epsfig}
\usepackage{graphicx}
\usepackage{bbm}
\usepackage{wrapfig}

\usepackage{tikz}
\usepackage{comment}

\usepackage{algorithm}
\usepackage{algorithmic}

\usepackage{caption}
\usepackage{subcaption}

\usepackage[bb=dsserif]{mathalpha}
\usepackage{bm}

\usepackage{booktabs,colortbl,tabularx}
\usepackage{pifont}%

\definecolor{Gray}{gray}{0.9}

\newcommand{\cmark}{\ding{51}}%
\newcommand{\xmark}{\ding{55}}%

\definecolor{citecolor}{HTML}{2980b9}
\definecolor{linkcolor}{HTML}{c0392b}
\usepackage[pagebackref=true,breaklinks=true,colorlinks,bookmarks=false,citecolor=citecolor,linkcolor=linkcolor]{hyperref}

\title{Unified Video-Language Pre-training with Synchronized Audio}

\author{%
  Shentong Mo$^{1,2*}$~Haofan Wang$^{3}$~Huaxia Li$^{3}$~Xu Tang$^{3}$\\
  $^1$CMU, $^2$MBZUAI, $^3$Xiaohongshu
}

\begin{document}

\maketitle

\begin{abstract}

Video-language pre-training is a typical and challenging problem that aims at learning visual and textual representations from large-scale data in a self-supervised way.
Existing pre-training approaches either captured the correspondence of image-text pairs or utilized temporal ordering of frames.
However, they do not explicitly explore the natural synchronization between audio and the other two modalities.
In this work, we propose an enhanced framework for Video-Language pre-training with Synchronized Audio, termed as VLSA, that can learn tri-modal representations in a unified self-supervised transformer.
Specifically, our VLSA jointly aggregates embeddings of local patches and global tokens for video, text, and audio.
Furthermore, we utilize local-patch masked modeling to learn modality-aware features, and leverage global audio matching to capture audio-guided features for video and text.
We conduct extensive experiments on retrieval across text, video, and audio.
Our simple model pre-trained on only 0.9M data achieves improving results against state-of-the-art baselines.
In addition, qualitative visualizations vividly showcase the superiority of our VLSA in learning discriminative visual-textual representations.

\end{abstract}

\section{Introduction}

When we enjoy fascinating story in a movie, not only frames with the caption are attractive, but synchronized sounds are also impressive for us.
In daily life, sound appears to be in multiple scenes, such as a classroom scene with professors’ speech and students' whispering, a family reunion that consists of talking and laughing.
As audio and visual contents are commonly matched and synchronized, many researchers~\cite{mo2022benchmarking,mo2022semantic,mo2023diffava,mo2023class,pian2023audiovisual,mo2023oneavm,mo2024texttoaudio} have explored the benefit of such synchronization for audio-visual tasks, such as audio-event localization~\cite{tian2018ave,lin2019dual,wu2019dual,lin2020audiovisual,mo2022multimodal,mo2023deepavfusion}, audio-visual spatialization~\cite{Morgado2018selfsupervised,gao20192.5D,Chen2020SoundSpacesAN,Morgado2020learning}, and sound source localization~\cite{Senocak2018learning,hu2019deep,Afouras2020selfsupervised,qian2020multiple,chen2021localizing,mo2022EZVSL,mo2022slavc,mo2023audiovisual,mo2023avsam,mo2023weaklysupervised}.
However, in this work, we leverage this cross-modal synchronization between audio and videos/texts for enhancing video-language pre-training.

\begin{figure}[t]
\centering
\includegraphics[width=0.9\linewidth]{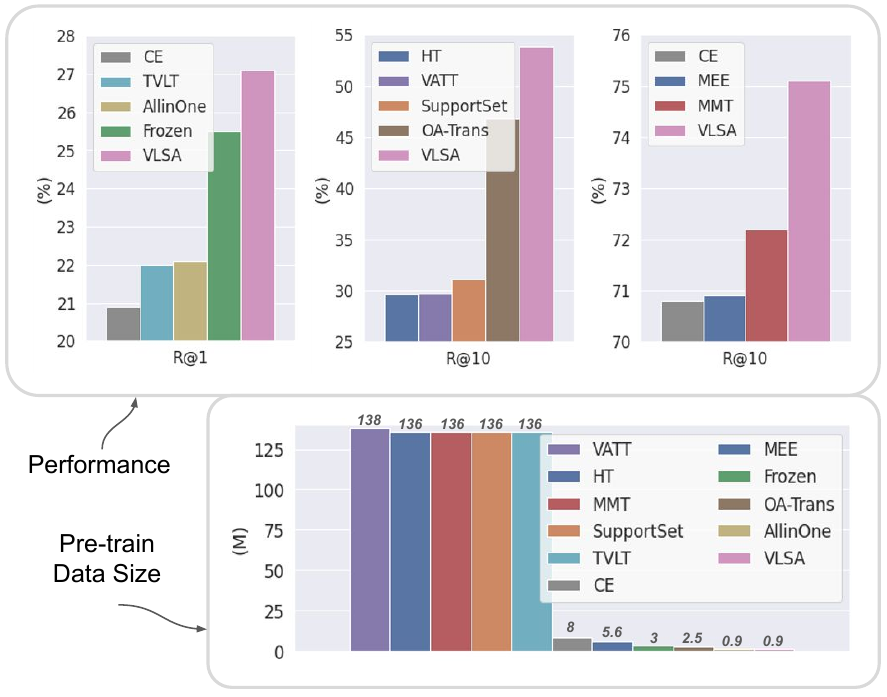}
\vspace{-0.7em}
\caption{{\bf Comparison performance of text-to-video (Left), zero-shot text-to-video (Middle), and text-to-audio retrieval (Right).}
VLSA pre-trained on only 0.9M data achieves significant gains compared to previous video-language (MEE, HT, MMT, SupportSet, Frozen, OA-Trans, AllinOne), 
video-audio (TVLT),
and video-language-audio (CE, VATT) methods.}
\label{fig: title_img}
\end{figure}

Video-language pre-training aims to learn visual and textual representations jointly from large-scale data.
Previous work~\cite{Liu2019a,gabeur2020mmt,Zhu2020ActBERTLG,sun2019videobert,Bain2021frozen,cheng2021improving,wang2022oatrans,wang2022allinone} either proposed to learn the correspondence of image-text pairs or utilized temporal ordering of frames.
Typically, MMT~\cite{gabeur2020mmt} proposed a multi-modal transformer to aggregate per-frame visual features with temporal information.
With the success of CLIP~\cite{radford2021learning} on visual and textual representation learning, CLIP2TV~\cite{Gao2021CLIP2TVAE} leveraged CLIP pre-trained weights with a video-text alignment module and a video-text matching module for discriminating positive and negative pairs of embeddings from text and video encoders.

However, they do not explicitly explore the natural synchronization between audio and other two modalities, and can not learn compact representations of the global correspondence across each modality.
In contrast, we leverage synchronized audio for global audio matching in our unified transformer to capture discriminative global features for boosting video-language pre-training.

Since audio and visual contents are commonly matched and synchronized, recent pre-training methods~\cite{akbari2021vatt, guzhov2021audioclip,zellers2022merlot} are developed to incorporate audio as an auxiliary modality for learning discriminative representations transferrable to video-text retrieval.
VATT~\cite{akbari2021vatt} introduced a multi-modal contrastive loss for global features of each modality to learn the alignment of video-audio-text triplets.
AudioCLIP~\cite{guzhov2021audioclip} extended CLIP with the ESResNeXt audio model and optimized a contrastive loss between two modalities to maximize diagonal values of the scaled dot-product similarity matrix.
More recently, MERLOT Reserve~\cite{zellers2022merlot} employed a contrastive span training loss between masked tokens and corresponding global uni-modal embeddings for audio, captions, and video frames.
While these methods achieve promising results on video-language pre-training, they are extremely dependent on the capacity of three separate encoders with many parameters to extract discriminative representations for each modality on large-scale datasets.
In addition, they do not incorporate local features into the pre-training objective.
Different from them, we combine local-patch masked modeling with global-token alignment in a unified transformer to learn modality-aware features during pre-training.

To this end, we propose a novel and enhanced framework for Video-Language pre-training with Synchronized Audio, namely VLSA, that can jointly learn tri-modal representations in a unified self-supervised transformer.
Specifically, VLSA aggregates local patches and global token representations for each modality through a joint transformer.
Furthermore, local-patch masked modeling is applied to patch-level embeddings of each modality to learn modality-aware local features.
In addition, global audio matching is applied on both video and text modalities to capture compact features from global tokens of audio, text, and video frames.

We pre-train our VLSA on only 0.9M video-audio-transcript triplets from an intersection set of HowTo100M and AudioSet.
We conduct extensive experiments on retrieval across text, video, and audio. 
Our simple pre-trained model achieves improving results against previous baselines on downstream tasks.
In addition, qualitative visualizations vividly showcase the advantage of the proposed approach in learning compact visual-textual representations.

Our contributions can be summarized as follows:
\begin{itemize}
    \item We propose a novel and enhanced framework for Video-Language pre-training with Synchronized Audio, termed as VLSA, that aggregates local patches and global visual-textual representation guided by audio  through a joint transformer encoder.
    \item We introduce audio local-patch masked modeling to learn modality-aware interaction between audio and the other two modalities.
    \item We further leverage global audio matching to capture compact and discriminative video-text features without the need for video-text matching.
    \item Our simple unified transformer pre-trained on only 0.9M triplets achieves competitive results against baselines on text-video and text-audio retrieval.
    
\end{itemize}

\vspace{-0.5em}
\section{Related Work}

\noindent \textbf{Video-Language Pre-training.}
Video-language pre-training aims at learning joint visual and textual representations from captions and video frames.
At the early stage, CE~\cite{Liu2019a} designed a collaborative experts model to aggregate information from different pre-trained experts for video retrieval tasks.
MEE~\cite{Miech2018LearningAT} proposed a model with mixed embedding experts to handle missing input modalities for learning improved text-video embeddings simultaneously.
In the recent years, diverse pipelines~\cite{imagebert,kvlbert,ernievil,uniter,sun2019videobert,tan2019lxmert,li2019visual,lu2019vilbert,Su2020VL-BERT:,shi2020contrastive,hong2021recurrent,lei2021less,Bain2021frozen,wang2022oatrans,wang2022allinone,ge2022miles} have been proposed to explore the fusion of two distinct modalities, where the correspondence of image-text pairs is usually aligned during pre-training, as shown in Table~\ref{tab: summary_work}.
Typically, ActBERT~\cite{Zhu2020ActBERTLG} leveraged a tangled transformer to learn the correspondence between global actions and local object regions from paired video sequences and text descriptions.
A multi-modal transformer was introduced in MMT~\cite{gabeur2020mmt} to aggregate per-frame visual features with temporal information.
CAMoE~\cite{cheng2021improving} utilized mixture-of-experts to capture the alignment between text and multiple video representations of action and scene.
Based on CLIP~\cite{radford2021learning} pre-trained weights, CLIP2TV~\cite{Gao2021CLIP2TVAE} adopted a video-text alignment module and a video-text matching module to discriminate positive and negative pairs of features from text and video encoders.

While the aforementioned video-language pre-training approaches achieve promising performance on downstream tasks, they do not learn the natural synchronization between audio and the other two modalities in an explicit manner.
Moreover, they can not learn compact representations of the global correspondence across each modality for retrieval tasks~\cite{Oncescu21a,Koepke2022audio} involved with audio.
In this work, we aim to enhance video-language pre-training with synchronized audio by global-token matching for both text-video and text-audio retrieval.

\begin{table*}[t]
        \renewcommand\tabcolsep{11.0pt}
	\centering
        \vspace{-0.5em}
        \caption{{\bf Comparison of multi-modal self-supervised pre-training on video, text, and audio or speech.}
	``Single'', ``Dual'', and ``Triple'' refer to one joint encoder, two separate encoders, and three separate encoders, respectively.
	Local and global denote uni-modal masked modeling and multi-modal matching. }
	\label{tab: summary_work}
	\renewcommand\tabcolsep{4.0pt}
	\scalebox{0.75}{
		\begin{tabular}{lccccc}
		    \toprule
		    Method & Publication & Modalities & Architecture & Pre-train Objectives & Pre-train Datasets \\
		    \midrule
		    VideoBERT~\cite{sun2019videobert} &ICCV, 2019 &  Video, Text & Single & local & self-collected \\
		    Frozen~\cite{Bain2021frozen} & ICCV, 2021 &  Video, Text & Dual & global & WebVid2M+CC3M \\
		    OA-Trans~\cite{wang2022oatrans} & CVPR, 2022 & Video, Text & Dual & global & WebVid2M+CC3M \\
                Region-Learner~\cite{yan2021video} & AAAI, 2023 & Video, Text & Single & global & WebVid2M+CC3M \\
		    AllinOne~\cite{wang2022allinone} & CVPR, 2023 & Video, Text & Single & local+global & WebVid2M+HowTo100M \\ \hline
		    VATT~\cite{akbari2021vatt} & NeurIPS, 2021 & Video, Text, Audio & Single/Triple & global & AudioSet+HowTo100M \\
		    AudioCLIP~\cite{guzhov2021audioclip} & ICASSP, 2022 & Video, Text, Audio & Triple & global & AudioSet \\
		    MERLOT Reserve~\cite{zellers2022merlot} & CVPR, 2022 & Video, Text, Audio & Triple & global & YT-Temporal-1B \\ 
		    i-Code~\cite{Yang2022iCodeAI} & AAAI, 2023 & Video, Text, Speech & Single & local+global & self-collected \\ 
                \rowcolor{gray!20}
		    VLSA (ours) & -- & Video, Text, Audio & Single & local+global & AudioSet$\cap$HowTo100M  \\ 
			\bottomrule
			\end{tabular}}
 \vspace{-1.0em}
\end{table*}

\noindent \textbf{Audio Synchronization in Pre-training.}
Audio synchronization in pre-training has been applied in previous works~\cite{akbari2021vatt,guzhov2021audioclip,zellers2022merlot,Yang2022iCodeAI} to learn discriminative representations of audio, captions, and video frames.
VATT~\cite{akbari2021vatt} introduced a multi-modal contrastive loss for global features of each modality to learn the alignment of video-audio-text triplets.
AudioCLIP~\cite{guzhov2021audioclip} extended CLIP with the ESResNeXt audio model and optimized a contrastive loss between two modalities to maximize diagonal values of the scaled dot-product similarity matrix.
More recently, MERLOT Reserve~\cite{zellers2022merlot} employed a contrastive span training loss between masked tokens and corresponding global uni-modal embeddings for audio, captions, and video frames.
i-Code~\cite{Yang2022iCodeAI} exploited a multi-modal fusion network to integrate single-modality outputs of vision, speech, and language from each pre-trained encoder.

Different from these tri-modal pre-training baselines, the proposed transformer is more lightweight than comparable methods, as we do not need to use three separate encoders with many parameters to extract discriminative embeddings for each modality.
In addition, we develop a fully novel and unified transformer to aggregate both local-patch masked modeling and global-token matching for learning modality-aware features.
Audio-video matching and masked modeling are also addressed in a very recent work called TVLT~\cite{tang2022tvlt}, but they do not involve captions during pre-training.

\vspace{-0.5em}
\section{Method}

Given a triplet of video sequences, captions, and synchronized audio, our target is to learn joint representations from large-scale unsupervised data. 
We propose a novel Video-Language pre-training framework with Synchronized Audio named VLSA for enhancing visual-textual semantics from sound itself, which mainly consists of two modules, Local-Patch Masked Modeling in Section~\ref{sec:lpmm} and Global Audio Matching in Section~\ref{sec:gam}.

\begin{figure*}[t]
    \centering
    \includegraphics[width=0.98\linewidth]{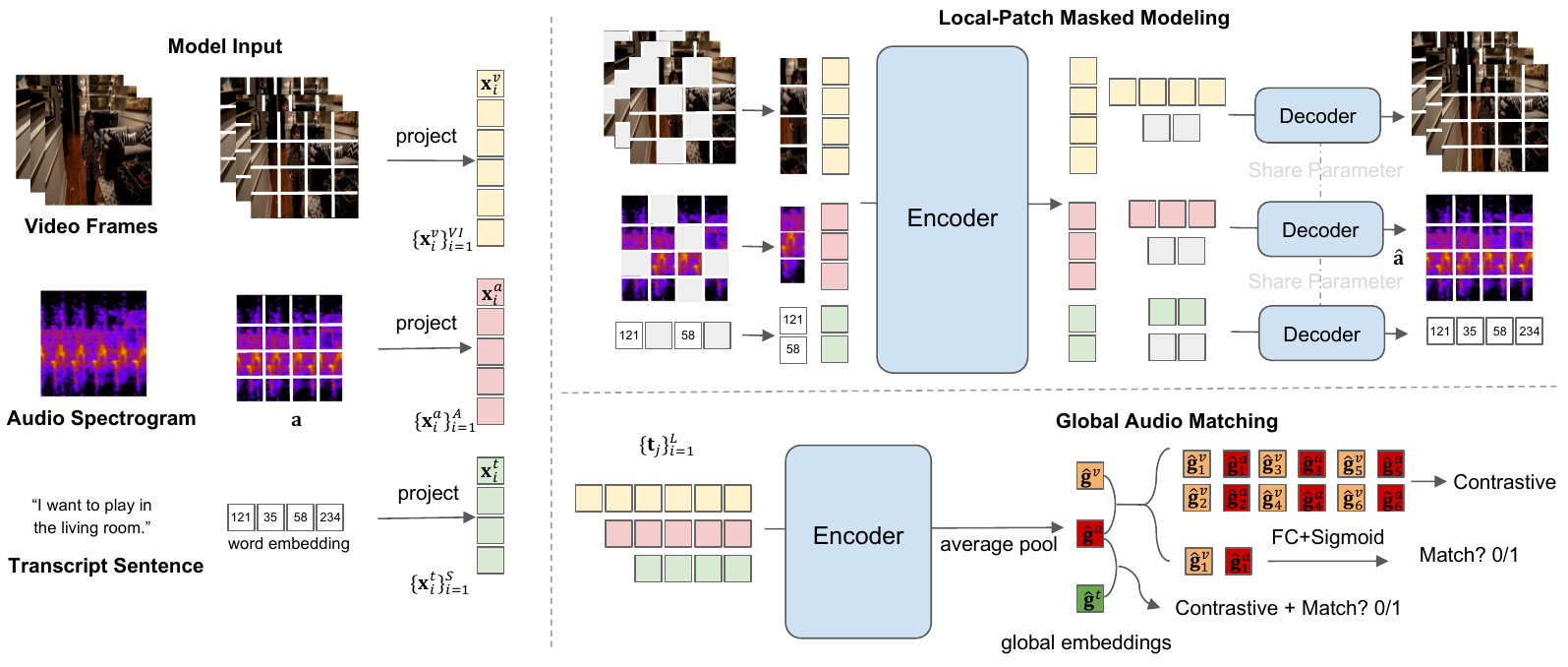}
    \vspace{-0.5em}
    \caption{{\bf Illustration of our enhanced framework of Video-Language pre-training with Synchronized Audio (VLSA).}
    The modality-aware patch embeddings $\{\mathbf{x}_i^v\}_{i=1}^{VI}$, $\{\mathbf{x}_i^a\}_{i=1}^A$, $\{\mathbf{x}_i^t\}_{i=1}^S$, are extracted from each linear projection layer.
    The Local-Patch Masked Modeling module is applied to local-patch representations for audio spectrogram $\mathbf{a}$ extracted from the unified encoder, and the decoder is utilized to predict the raw audio spectrograms $\hat{\mathbf{a}}$ for learning the interaction of audio and the other two modalities (video and text).
    Finally, the Global Audio Matching module (contrastive loss and binary matching loss) is leveraged on modality-aware global embeddings $\hat{\mathbf{g}}^v, \hat{\mathbf{g}}^t, \hat{\mathbf{g}}^a$ averaged from the encoder to capture the cross-modal alignment between synchronized audio and video frames/caption sentence in an explicit manner.
    }
	\label{fig: main_img}
\end{figure*}

\subsection{Preliminaries}

In this section, we first describe the problem setup and notations, and then revisit the commonly-used objective for video-language pre-training.

\noindent\textbf{Problem Setup and Notations.}
Given a triplet of video sequences with a dimension of $V\times 3\times H\times W$, texts with a dimension of $S\times C$, and synchronized audio spectrograms with a dimension of $T\times F$, our target is to learn discriminative representations simultaneously and evaluate them for downstream retrieval tasks.
We formally denote $V$ as the number of video frames, $S$ as the length of text sentences, $C$ as the length of the word dictionary, and $F$ as the number of frequencies in audio. 
$H$ and $W$ are the height and width of each frame in the video.

\noindent\textbf{Revisit Video-Language Pre-training.}
To solve the joint modeling problem, current pre-training baselines~\cite{li2019visual,lu2019vilbert,Su2020VL-BERT:,shi2020contrastive,hong2021recurrent,lei2021less,Bain2021frozen,wang2022oatrans,wang2022allinone} introduced a visual-textual contrastive loss to learn the alignment between images and captions.
Given a set of global video-language features $\{\mathbf{g}^v_i\}_{i=1}^B,\{\mathbf{g}^t_i\}_{i=1}^B$ in a batch of size $B$,
the contrastive loss between the cosine similarity $\mathtt{sim}(\mathbf{g}^v_i, \mathbf{g}^t_i)$ of global embeddings is formulated as:
\begin{equation}\label{eq:vt_cl}
    \mathcal{L}_{v\rightarrow t} = \dfrac{1}{B}\sum_i^B -\log \dfrac{\exp\left(\frac{1}{\tau}\mathtt{sim}(\mathbf{g}^v_i, \mathbf{g}^t_i)\right)}{\sum_{j=1}^B \exp\left(\frac{1}{\tau}\mathtt{sim}(\mathbf{g}^v_i, \mathbf{g}^t_j)\right)},
\end{equation}
where $\mathbf{g}^v_i, \mathbf{g}^t_j\in\mathbb{R}^{1\times D}$, and $D$ is the size of the embedding dimension. 
$\mathtt{sim}(\mathbf{g}^v_i, \mathbf{g}^t_i)=(\mathbf{g}^v_i)^\top\mathbf{g}^t_i / ( \|\mathbf{g}^v_i\| \|\mathbf{g}^t_i\|)$ is the cosine similarity, and $\tau$ is the temperature parameter.
$B^2-B$ negative vision-language pairs are created within a training batch.
Besides, they proposed to take $N^{v,t}$ samples randomly from the whole dataset and applied an FC layer and sigmoid operator to predict the matching probability $\mathbf{p}^{v,t}$ of image-caption pairs. 
The video-text matching loss is formally summarized as:
\begin{equation}\label{eq:vtm}
    \mathcal{L}_{vtm} = \sum_{n=1}^{N^{v,t}}\mbox{BCE}\left(\mathbf{y}^{v,t}_{n}, \mathbf{p}^{v,t}_{n}\right)
\end{equation}
where $\mathbf{y}^{v,t}_{n}, \mathbf{p}^{v,t}_{n}$ denote the ground-truth and probability of $n$th visual-textual representation pairs.
$\mbox{BCE}(\cdot)$ is a binary cross-entropy loss.
If the image and caption are from the same pair, the target is 1; otherwise, it is 0.
During training, the positive pair is sampled from the dataset, while the negative pair is created by replacing the text or vision in a paired sample with a randomly selected other sample.
Overall, the global loss for video-language alignment can be computed by $\mathcal{L}_{v,t} = \mathcal{L}_{v\rightarrow t} + \mathcal{L}_{t\rightarrow v} + \mathcal{L}_{vtm}$.

However, those video-language pre-training approaches are extremely dependent on the capacity of encoders with many parameters to extract discriminative global embeddings for each modality.
In addition, most frameworks do not explicitly learn the natural synchronization between audio and video/text.
To address this issue, we propose a novel and unified transformer for Video-Language pre-training with Synchronized Audio, that can learn both local and global features between audio and videos/captions simultaneously, as illustrated in Figure~\ref{fig: main_img}.

\subsection{Local-Patch Masked Modeling}\label{sec:lpmm}
In order to learn local modality-aware features across three different modalities, we introduce modality-aware patch embeddings for each modality that are extracted from raw input via each linear projection layer, \textit{i.e.}, $\mathbf{x}^v\in\mathbb{R}^{(V\times I)\times D}$,
$\mathbf{x}^t\in\mathbb{R}^{S\times D}$,
$\mathbf{x}^a\in\mathbb{R}^{A\times D}$, where $I, S, A$ denotes the total number of patches for each video frame, each caption, and the corresponding audio.
Assume the patch resolution of each frame and audio are $P^v, P^a$, the patch-wise raw input for video and audio are formally denoted as $\mathbf{v}\in\mathbb{R}^{(V\times I) \times (3\times P^v\times P^v)}$ and $\mathbf{a}\in\mathbb{R}^{A\times (P^a\times P^a)}$.
Note that $I = H/P^v\times W/P^v, A = T/P^a\times F/P^a$.

With local-patch representations for each modality $\{\mathbf{x}^v_i\}_{i=1}^{VI}, \{\mathbf{x}^t_i\}_{i=1}^{S}, \{\mathbf{x}^a_i\}_{i=1}^{A}$, we first apply an unified transformer encoder $\phi(\cdot)$ to aggregate patch-level features from the raw input as:
\begin{equation}\label{eq:loc}
\begin{aligned}
    \{\hat{\mathbf{x}}^v\}_{i=1}^{VI}, \{\hat{\mathbf{x}}^t_i\}_{i=1}^S,
    \{\hat{\mathbf{x}}^a_i\}_{i=1}^A = \{\phi(\mathbf{x}_j, \mathbf{X}, \mathbf{X})\}_{j=1}^{VI+S+A}, \\ 
    \mathbf{X} = \{\mathbf{x}_j\}_{j=1}^{VI+S+A} = [\{\mathbf{x}^v_{i}\}_{i=1}^{VI}; \{\mathbf{x}^t_i\}_{i=1}^S; \{\mathbf{x}^a_i\}_{i=1}^A]
\end{aligned}
\end{equation}
where $[\ ;\ ]$ denotes the concatenation operator.
$\mathbf{x}^v_i,\mathbf{x}^t_i,\mathbf{x}^a_i\in\mathbb{R}^{1\times D}$, and $D$ is the dimension of embeddings.
The self-attention operator $\phi(\cdot)$ is formulated as:
\begin{equation}
    \phi(\mathbf{x}_j, \mathbf{X}, \mathbf{X}) = \mbox{Softmax}(\dfrac{\mathbf{x}_j\mathbf{X}^\top}{\sqrt{D}})\mathbf{X}
\end{equation}
Then, to capture the interaction across each modality, we exploit a tri-modal masking mechanism and a shared decoder to predict the masked patch of one modality, given the other two modalities as auxiliaries.
Specifically, with video and text embeddings $\{\mathbf{x}^v_i\}_{i=1}^{VI}, \{\mathbf{x}^t_i\}_{i=1}^{S}$, we leverage a decoder to predict the raw audio spectrograms $\hat{\mathbf{a}}$ for randomly masked audio patches.
The local-level audio masked loss is computed with the mean square loss between the targeted and predicted spectrograms as:  
\begin{equation}
    \mathcal{L}^{\text{local}}_a = \dfrac{1}{N^a}\sum_{i\in M^a} ||\mathbf{a}_i - \hat{\mathbf{a}}_i||_2^2
\end{equation}
where $N^a, M^a$ denotes the total number and the set of masked patches for audio, respectively.
Similar to audio, the local-level video masked loss $\mathcal{L}^{\text{local}}_v$ is calculated with the mean square loss between the ground truth and missing pixel of frames with a random mask on all $VI$ patches.
For text masked loss $\mathcal{L}^{\text{local}}_t$, we use a decoder to predict the randomly masked token in $S$ tokens, similar to BERT~\cite{devlin2018bert}.
Note that three separate decoders for tri-modal masked prediction are parameter-shared and achieve the best performance, as observed in our experiments in Section~\ref{sec: exp_ab}.
With optimizing the total local-level masked loss $\mathcal{L}^{\text{local}}=\sum_{m\in\{a,v,t\}}\mathcal{L}^{\text{local}}_m$ with a shared encoder, it will capture the local-level interaction between audio and other two distinct modalities, which pushes the model to learn more discriminative embeddings.

\subsection{Global Audio Matching}\label{sec:gam}

Benefiting from the local-level masked loss above, we propose a novel and explicit global audio matching mechanism on global embeddings $\hat{\mathbf{g}}^v, \hat{\mathbf{g}}^t, \hat{\mathbf{g}}^a \in\mathbb{R}^{1\times D}$ for each modality in the unified transformer $\phi(\cdot)$ to generate global modality-aware features as:
\begin{equation}
\begin{aligned}
\hat{\mathbf{g}}^v;\hat{\mathbf{g}}^t;\hat{\mathbf{g}}^a & = \mbox{AvgPool}(\{\hat{\mathbf{x}}^v\}_{i=1}^{VI};\{\hat{\mathbf{x}}^t_i\}_{i=1}^S;\{\hat{\mathbf{x}}^a_i\}_{i=1}^A); \\
\end{aligned}
\end{equation}
where $\hat{\mathbf{g}}^v, \hat{\mathbf{g}}^t, \hat{\mathbf{g}}^a\in\mathbb{R}^{1\times D}$, and $D$ is the dimension of each global embedding.
$\mbox{AvgPool}(\cdot)$ denotes the average pooling operator.
That is, we average local-level patches for three modalities along each total number of patches ($VI$, $S$, $A$) to generate the modality-aware global embeddings.
$[\ ;\ ]$ is the concatenation operator.

In order to explicitly learn the cross-modal alignment between synchronized audio and video frames, we leverage a contrastive loss similar to Eq.~\ref{eq:vt_cl} for maximizing the cosine similarity $\mathtt{sim}(\hat{\mathbf{g}}^a_i,\hat{\mathbf{g}}^v_i)$ of audio-video pairs from the same batch index $i$.
By applying an FC layer and sigmoid operator to predict the alignment probability $\mathbf{p}^{a,v}\in\mathbb{R}^{N\times 1}$ of $N^{a,v}$ audio-video pairs that are randomly chosen from the whole pre-training dataset, we formulate the global audio-visual alignment loss as:
\begin{equation}\label{eq:avm}
\begin{aligned}
    & \mathcal{L}_{a\rightarrow v}^{\text{global}} = \dfrac{1}{B}\sum_i^B -\log \dfrac{\exp\left(\frac{1}{\tau}\mathtt{sim}(\hat{\mathbf{g}}^a_i, \hat{\mathbf{g}}^v_i)\right)}{\sum_{j=1}^B \exp\left(\frac{1}{\tau}\mathtt{sim}(\hat{\mathbf{g}}^a_i, \hat{\mathbf{g}}^v_j)\right)} \\
    & -\log \dfrac{\exp\left(\frac{1}{\tau}\mathtt{sim}(\hat{\mathbf{g}}^v_i, \hat{\mathbf{g}}^a_i)\right)}{\sum_{j=1}^B \exp\left(\frac{1}{\tau}\mathtt{sim}(\hat{\mathbf{g}}^v_i, \hat{\mathbf{g}}^a_j)\right)} + \sum_{n=1}^{N^{a,v}}\mbox{BCE}\left(\mathbf{y}^{a,v}_{n}, \mathbf{p}^{a,v}_{n}\right) \\
\end{aligned}
\end{equation}
where $B$ is the batch size.
$\mathbf{y}^{a,v}\in\mathbb{R}^{N\times 1}$ denotes a one-hot encoding and its element for the entry is 0 for non-alignment and 1 for alignment.
Since one audio spectrogram is distinct in each audio-video pair, this alignment loss does not bring false alignment pairs, while most frames in a video look similar in high-level textual semantics.
To boost the compactness of pre-trained global representations for retrieval, we apply similar cross-modal alignment loss $\mathcal{L}_{a\rightarrow t}^{\text{global}}$ on global tokens across audio-text pairs.

With optimizing the total global-token alignment loss $\mathcal{L}^{\text{global}}=\mathcal{L}_{a\rightarrow v}^{\text{global}}+\mathcal{L}_{a\rightarrow t}^{\text{global}}$, we push the model to learn more discriminative video and text representations with the benefit of synchronized audio.
The overall objective of our model is simply optimized in an end-to-end manner as
\begin{equation}
    \mathcal{L} = \mathcal{L}^{\text{local}} + \lambda\cdot \mathcal{L}^{\text{global}}
\end{equation}
where $\lambda$ denotes the weighted parameter for balancing two losses with different orders of magnitude.
We use $\lambda = 5$ as the default in our experiments.
During inference, we simply compute the cosine-similarity $\mathtt{sim}(\hat{\mathbf{g}}^t,\hat{\mathbf{g}}^v),\mathtt{sim}(\hat{\mathbf{g}}^t,\hat{\mathbf{g}}^a), \mathtt{sim}(\hat{\mathbf{g}}^v,\hat{\mathbf{g}}^a)$ for retrieval across text-video, text-audio, and video-audio settings.

\begin{table*}[t]
    \renewcommand\tabcolsep{8.0pt}
    \renewcommand{\arraystretch}{1.1}
	\centering
        \vspace{-0.5em}
	\caption{{\bf Quantitative results of text-to-video retrieval on MSR-VTT dataset.} 
    ``Single'', ``Dual'', and ``Triple'' refer to one joint encoder, two separate, and three separate encoders.
    Bold and Underline denote the best and second result, respectively. }
	\label{tab: exp_msrvtt}
	\scalebox{0.63}{
	\begin{tabular}{lcccccccc}
			\toprule
			\bf Method & \bf Modalities & \bf Architecture & \bf Pre-train Datasets & \bf Data Size ($\downarrow$) & \bf R@1 ($\uparrow$) & \bf R@5 ($\uparrow$) & \bf R@10 ($\uparrow$) \\
			\midrule
		   JSFusion~\cite{Yu2018AJS} &  Video, Text & Dual & AudioSet+ImageNet & 3M & 10.2 & 31.2 & 43.2 \\
     MEE~\cite{Miech2018LearningAT} & Video, Text, Audio & Triple & COCO+VisGenome & 5.6M & 14.2 & 39.2 & 53.8 \\
		   HT~\cite{Miech2019howto100m} &  Video, Text & Dual & HowTo100M & 136M & 14.9 & 40.2 & 52.8 \\ 
     CE~\cite{Liu2019a} &  Video, Text, Audio & Triple & YouTube-8M & 8M & 20.9 & 48.8 & 62.4 \\ 
            AVLnet~\cite{rouditchenko2020avlnet} & Video, Text, Audio & Triple &  HowTo100M & 136M & \textbf{27.1} & 55.6 & 66.6 \\
     ActBERT~\cite{Zhu2020ActBERTLG} &  Video, Text & Dual & HowTo100M & 136M & 16.3 & 42.8 & 56.9 \\
		   HERO~\cite{li2020hero} &  Video, Text & Dual & HowTo100M & 136M & 16.8 & 43.4 & 57.7 \\
			VidTranslate~\cite{Korbar2020VideoUA} &  Video, Text & Dual & HowTo100M & 136M & 14.7 & - & 52.8 \\
			NoiseEstimation~\cite{Amrani2021NoiseEU}&  Video, Text & Dual & HowTo100M & 136M & 17.4 & 41.6 & 53.6 \\
			UniVL~\cite{Luo2020UniVL} &  Video, Text & Dual & HowTo100M & 136M & 21.2 & 49.6 & 63.1\\
			
			MMT~\cite{gabeur2020mmt} &  Video, Text, Audio & Triple & HowTo100M & 136M & 26.6 & \underline{57.1} & \textbf{69.6} \\ 
			ClipBERT~\cite{lei2021less} &  Video, Text & Dual & COCO+VisGenome & 5.6M &  22.0 & 46.8 & 59.9 \\
            Frozen~\cite{Bain2021frozen} &  Video, Text & Single & CC3M & 3M & 25.5 & 54.5 & 66.1 \\
            TVLT~\cite{tang2022tvlt} & Video, Audio & Single & HowTo100M & 136M & 22.0 & - & - \\ 
            Everything-At-Once~\cite{shvetsova2022everything} & Video, Text, Audio & Triple &  HowTo100M & 136M & 23.7 & 52.1 & 63.7 \\
            AllinOne~\cite{wang2022allinone} &  Video, Text & Single &
            AudioSet$\cap$HowTo100M & \textbf{0.9M} & 22.1& 49.1 & 60.6 \\
            \rowcolor{gray!20}
		VLSA (ours) & Video, Text, Audio & Single &  AudioSet$\cap$HowTo100M & \bf 0.9M & \bf 27.1 & \bf 57.3 & \underline{68.9} \\ \hline
			\textit{zero-shot:} \\
			HT~\cite{Miech2019howto100m} & Video, Text & Dual &  HowTo100M & 136M & 7.5 & 21.2 & 29.6\\
			SupportSet~\cite{patrick2021supportset} & Video, Text & Dual & HowTo100M & 136M & 8.7 & 23.0 & 31.1 \\
		VATT~\cite{akbari2021vatt} & Video, Text, Audio & Single &  AudioSet+HowTo100M & 138M & -- & -- & 29.7 \\
  Frozen~\cite{Bain2021frozen} & Video, Text & Dual & CC3M+WebVid-2M & 5.5M & \underline{18.7} & \underline{39.5} & \underline{51.6} \\
			OA-Trans~\cite{wang2022oatrans} & Video, Text & Dual & WebVid-2M & 2.5M & 18.4 & 36.5 & 46.8\\ 
            Everything-At-Once~\cite{shvetsova2022everything} & Video, Text, Audio & Triple &  HowTo100M & 136M & 9.9 & 24.0 & 32.6 \\
            \rowcolor{gray!20}
		VLSA (ours) & Video, Text, Audio & Single & AudioSet$\cap$HowTo100M & \bf 0.9M & \bf 20.4 & \bf 40.6 & \bf 53.8 \\
			\bottomrule
			\end{tabular}}
\end{table*}

\section{Experiments}

\subsection{Experimental Setup}

\noindent\textbf{Datasets.}
HowTo100M~\cite{Miech2019howto100m} consists of 136M video clips from 1.22M YouTube videos with 134,472 hours.
AudioSet~\cite{gemmeke2017audioset} contains 2,084,320 clips with 632 classes from YouTube videos covering human and animal sounds, musical instruments, and common everyday environmental sounds.
An intersection of HowTo100M and AudioSet with 0.9M audio-video-text triplets is used for pre-training.
MSR-VTT~\cite{xu2016msrvtt} includes 10K YouTube videos with 200K description sentences and is split into 9K for training and 1K for testing.
LSMDC~\cite{Rohrbach2015a} contains 118,081 video clips with 7,408 and 1,000 videos for validation and testing.
AudioCaps~\cite{kim2019audiocaps} is filtered out for 49,291 clips for training, 428 clips for validation, and 816 clips for testing.
SoundDescs~\cite{Koepke2022audio} consists of 32,979 audio clips with 23 sound categories, and is divided into 70\% of the clips for training and 15\% each for validation and testing.
Yoocook2~\cite{zhou2018towards} comprises 13K video clips of 89 cooking recipes with 9,586 clips for training and 3,350 clips for validation.

\noindent\textbf{Evaluation Metrics.}
We use the standard retrieval metrics~\cite{Zhu2020ActBERTLG,lei2021less,Korbar2020VideoUA,Amrani2021NoiseEU,akbari2021vatt} to evaluate the performance of our model.
Recall at rank $k$ (R@$k$) measures the percentage of labels retrieved within the top $k$ ranked predictions, and the higher value is better.
$k$ = 1, 5, 10 for text-video retrieval, and $k$ = 1, 5, 10, 50 for text-audio retrieval.
For all metrics, we report the average result of three different random seeds.

\noindent\textbf{Implementation.}
For each audio waveform, we follow the prior work~\cite{zhao2018the} and sub-sample the audio signal to 11kHz.
A Short-time Fourier transform with a window size of 1022 and a hop length of 256 is further applied to generate a $512\times256$ Time-Frequency representation of the audio, which is resampled to a log-frequency scale with a size of $256\times256$ as the input audio spectrogram, \textit{i.e.}, $T=256, F=256$.
For each caption, the tokenizer embedding from BERT~\cite{devlin2018bert} is used as the input with a maximum sentence length of 40 and a vocab size of 30,522, \textit{i.e.}, $C=30,522$.
For each video clip, we randomly sample 8 frames as inputs and resize each frame to $224\times224$, \textit{i.e.}, $V=8, H=W=224$.
Following prior work~\cite{he2021masked,Baade2022MAEAST}, we apply a patch size of $16\times16$ for both audio and video frames.
With patch size $P^v=16, P^a=16$, the total number of visual and audio patches $I=196, A=256$.
We use a ViT-base~\cite{dosovitskiy2020image} model for the masked autoencoder same as in MAE~\cite{he2021masked}.
We follow previous approaches~\cite{devlin2018bert,he2021masked} to mask 15\% on each word token randomly, 75\% on patches of each frame independently, and 75\% on patches of audio spectrograms.
The model is trained for 200k steps with the AdamW~\cite{loshchilov2018decoupled} optimizer with a learning rate of 1e-4, a decay rate of 0.01, and a batch size of 2048.
For fine-tuning, the model is trained for 20 epochs with a batch size of 256.

\subsection{Comparison to Prior Work}

In this work, we propose a novel and effective framework for video-language pre-training with synchronized audio.
In order to validate the effectiveness of the proposed VLSA, we comprehensively compare it to previous video-text (VT), video-audio (VA), and video-text-audio (VTA) pre-training baselines:
i) \textbf{VT:} 
1) JSFusion~\cite{Yu2018AJS}: a very early sequence fusion model with multi-modal matching;
2) MEE~\cite{Miech2018LearningAT}: a baseline with mixed embedding experts for visual-textual modalities;
3) HT~\cite{Miech2019howto100m}: a simple baseline with two uni-modal encoders by learning text-video joint embedding;
4) ActBERT~\cite{Zhu2020ActBERTLG}: a tangled transformer with the global-local correlation between actions and object regions;
5) HERO~\cite{li2020hero}: a hierarchical architecture combined with cross-modal and temporal transformer;
6) VidTranslate~\cite{Korbar2020VideoUA}: a generative method with a translation objective between modalities;
7) NoiseEstimation~\cite{Amrani2021NoiseEU}: a multi-modal density estimation method for learning the cross-modal correspondence;
8) UniVL~\cite{Luo2020UniVL}: a unified pre-training model for video-language understanding and generation;
9) MMT~\cite{gabeur2020mmt}: a multi-modal transformer to aggregate per-frame visual features with temporal information;
10) ClipBERT~\cite{lei2021less}: a generic framework based on BERT with sparsely sampled short video clips for pre-training;
11) Frozen~\cite{Bain2021frozen}: a curriculum learning-based joint transformer with attention modeling on both space and time;
12) SupportSet~\cite{patrick2021supportset}: a generative method by recovering caption with a weighted combination of support visual features;
13) OA-Trans~\cite{wang2022oatrans}: an object-aware pre-training transformer with bounding boxes and tags as guidance;
14) AllinOne~\cite{wang2022allinone}: a unified transformer that learns joint representations from raw video and textual inputs.
ii) \textbf{VA:} 
15) TVLT~\cite{tang2022tvlt}: a very recent visual-audio pre-training framework with masked audio/video autoencoding and contrastive modeling to align video and audio.
iii) \textbf{VTA:} 
16) CE~\cite{Liu2019a}: a collaborative model with multiple pre-trained experts on each modality.
17) AVLnet~\cite{rouditchenko2020avlnet}: a self-supervised baseline that learns a shared audio-visual-textual embedding space directly from raw video and text signals;
18) VATT~\cite{akbari2021vatt}: a unified transformer with the multi-modal contrastive loss on global features of each modality for learning the alignment of video-audio-text triplets.

\begin{table*}[t]
    \renewcommand{\arraystretch}{1.0}
	\centering
        \vspace{-0.5em}
	\caption{{\bf Quantitative results of text-audio retrieval on SoundDescs dataset.} Best results are bold.}
	\label{tab: exp_soundescs}
	\scalebox{0.6}{
	\begin{tabular}{lccccccccccc}
			\toprule
			\multirow{2}{*}{\bf Method} & \multirow{2}{*}{\bf Pre-train Datasets} & \multirow{2}{*}{\bf Data Size ($\downarrow$)} & \multicolumn{4}{c}{\bf Text-to-Audio} & \multicolumn{4}{c}{\bf Audio-to-Text} \\
			& & & \bf R@1 ($\uparrow$) & \bf R@5 ($\uparrow$) & \bf R@10 ($\uparrow$)  & \bf R@50 ($\uparrow$) & \bf R@1 ($\uparrow$) & \bf R@5 ($\uparrow$) & \bf R@10 ($\uparrow$) & \bf R@50 ($\uparrow$)   \\
			\midrule
			MEE~\cite{Miech2018LearningAT} & YouTube-8M+VGGSound  & 8.2M & 30.8 & 60.8 & 70.9 & 85.9 & 30.9 & 60.3 & 70.1 & 85.3 \\
			CE~\cite{Liu2019a} & YouTube-8M+VGGSound  & 8.2M & 31.1 & 60.6 & 70.8 & 86.0 & 30.8 & 60.3 & 69.5 & 85.4 \\
			MMT~\cite{gabeur2020mmt} & YouTube-8M+VGGSound  & 8.2M & 30.7 & 61.8 & 72.2 & 88.8 & 31.4 & 63.2 & 73.4 & 89.0 \\
   \rowcolor{gray!20}
		VLSA (ours) & AudioSet$\cap$HowTo100M & \bf 0.9M & \bf 33.5 & \bf 63.7 & \bf 75.1 & \bf 91.6 & \bf 34.2 & \bf 65.9 & \bf 76.3 & \bf 92.1 \\ 
			\bottomrule
			\end{tabular}}
\end{table*}

\newcommand{\vtr}{
\begin{tabular}{lc}
    \toprule
    \bf Method & \bf R@1 \\
    \midrule
	 Frozen~\cite{Bain2021frozen} & 15.0 \\
  OA-Trans~\cite{wang2022oatrans} & 18.2 \\
  \rowcolor{gray!20}
  VLSA (ours) & \bf{19.3} \\
    \bottomrule
\end{tabular}
}

\newcommand{\atr}{
\begin{tabular}{lc}
    \toprule
    \bf Method & \bf R@1 \\
    \midrule
	 MEE~\cite{Miech2018LearningAT} & 26.6 \\
  MMT~\cite{gabeur2020mmt} & 39.6 \\
   \rowcolor{gray!20}
  VLSA (ours) & \bf{42.5} \\
    \bottomrule
\end{tabular}
}

\newcommand{\avr}{
\begin{tabular}{lc}
    \toprule
    \bf Method & \bf R@1 \\
    \midrule
	 AVLnet~\cite{rouditchenko2020avlnet} & 30.7 \\
  TVLT~\cite{tang2022tvlt} & 32.8 \\
   \rowcolor{gray!20}
  VLSA (ours) & \bf{36.3} \\
    \bottomrule
\end{tabular}
}

\begin{table*}[t]
    \centering
    \vspace{-0.5em}
    \caption{{\bf Quantitative results on LSMDC, AudioCaps, and Youcook2 benchmarks.}
    }
    \label{tab:exp_retrieval}
    \begin{subfigure}{0.26\textwidth}
        \resizebox{\linewidth}{!}{\vtr}
        \caption{Text-to-Video.}
        \label{tab:vtr}
    \end{subfigure}
    \begin{subfigure}{0.24\textwidth}
        \resizebox{\linewidth}{!}{\atr}
        \caption{Text-to-Audio.}
        \label{tab:atr}
    \end{subfigure}
    \begin{subfigure}{0.25\textwidth}
        \resizebox{\linewidth}{!}{\avr}
        \caption{Audio-to-Video.}
        \label{tab:avr}
    \end{subfigure}
    \vspace{-0.5em}
\end{table*}

\begin{table*}[t]
    \renewcommand{\arraystretch}{1.1}
	\centering
        \vspace{-0.5em}
	  \caption{{\bf Ablation studies on Local-Patch Masked Modeling (LPMM) and Global Audio Matching (GAM).}}
	\label{tab: ab_lpmm_gta}
	\scalebox{0.87}{
	    \begin{tabular}{cccccccc}
			\toprule
			\multirow{2}{*}{\bf LPMM} & \multirow{2}{*}{\bf GAM} & \multicolumn{3}{c}{\bf Text-to-Video} & \multicolumn{3}{c}{\bf Text-to-Audio} \\
			& & \bf R@1 & \bf R@5 & \bf R@10 & \bf R@1 & \bf R@5 & \bf R@10 \\
			\midrule
                \xmark & \xmark & 20.1 & 47.2 & 59.3 & 23.6 & 49.2 & 61.5 \\
			\cmark & \xmark & 22.6 & 49.7 & 61.2 & 28.5 & 53.6 & 66.3 \\
			\xmark & \cmark & 25.3 & 52.5 & 63.1 & 30.7 & 57.8 & 69.7 \\
   \rowcolor{gray!20}
			\cmark & \cmark & \textbf{27.1} & \textbf{57.3} & \textbf{68.9} & \textbf{33.5} & \textbf{63.7} & \textbf{75.1} \\ 
			\bottomrule
	    \end{tabular}}
\end{table*}

For text-video retrieval, we report the quantitative comparisons of fine-tuning and zero-shot results in Table~\ref{tab: exp_msrvtt}.
As can be seen, we achieve the best results in terms of R@1 and R@5 compared to all baselines fine-tuned on MSR-VTT.
In particular, the proposed VLSA significantly outperforms TVLT~\cite{tang2022tvlt}, the only video-audio pre-training approach with a single transformer, where we achieve the performance gains of 5.1 R@1.
In the meanwhile, we only need 0.9M data for pre-training to achieve competitive results compared to the performance (32.7 R@1, 60.9 R@5, and 72.5 R@10) of OA-Trans~\cite{wang2022oatrans}, which reduces 64\% of the least amount (2.5M) of pre-training data so far.
When pre-trained on the same amount of data, the proposed VLSA achieves performance gains of 5.0 R@1, 8.2 R@5, and 8.3 R@10, compared to AllinOne~\cite{wang2022allinone}, the unified video-language pre-training model.
These improvements demonstrate the effectiveness of our method in text-video retrieval by enhancing video-language pre-training with synchronized audio.

When compared to Frozen~\cite{wang2022oatrans}, the current state-of-the-art model pre-trained on 5.5M data, the proposed VLSA achieves zero-shot results gains of 1.7 R@1, 1.1 R@5, and 2.2 R@10.
Furthermore, our VLSA outperforms VATT~\cite{akbari2021vatt} by 24.1 R@10, which implies the importance of incorporating audio into local-patch masked modeling to learn discriminative modality-aware representations. 
Meanwhile, the proposed approach with a unified transformer pre-trained on 0.9M video-audio-text triplets still achieves this significant gain, compared to VATT~\cite{akbari2021vatt} pre-trained on 136M triplets and 2M video-audio pairs.
These results validate the superiority of our method in learning compact cross-modal embeddings for zero-shot text-video retrieval.

In addition, significant gains in text-audio retrieval can be observed in Table~\ref{tab: exp_soundescs}.
The proposed VLSA achieves the best performance in terms of all metrics for both text-to-audio and audio-to-text retrieval. 
When it comes to text-to-audio retrieval, our approach with a single transformer encoder obviously outperforms MMT with separate encoders pre-trained on 136M video-audio-text triplets by 2.8 R@1, 1.9 R@5, 2.9 R@10, and 2.8 R@50.
This further indicates the effectiveness of the proposed unified framework in learning discriminative audio and textual representations.
Results on more benchmarks are reported in Table~\ref{tab:exp_retrieval}.

\subsection{Experimental Analysis}\label{sec: exp_ab}


In this part, we performed ablation studies to demonstrate the benefit of introducing the Local-Patch Masked Modeling and Global Audio Matching modules.
In order to validate the effectiveness of local-patch masked modeling (LPMM) and global audio matching (GAM), we ablate the necessity of each module and report the quantitative results in Table~\ref{tab: ab_lpmm_gta}.
We can observe that adding LPMM highly improves the vanilla baseline without pre-training in terms of text-video (by 2.5 R@1, 2.5 R@5, and 1.9 R@10) and text-audio retrieval (by 1.9 R@1, 4.4 R@5, and 4.8 R10), which implies the benefit of LPMM in learning discriminative modality-aware embeddings.
Meanwhile, introducing only GAM in the baseline also increases the retrieval results by significant gains (5.2 R@1, 5.3 R@5, and 3.8 R@10 for text-video; 7.1 R@1, 8.6 R@5, and 8.2 R10 for text-audio).
More importantly, incorporating LPMM and GAM together highly raises the baseline by 7.0 R@1, 10.1 R@5, 9.6 R@10 for text-video, and 9.9 R@1, 14.5 R@5, 13.6 R10 for text-audio.
These results demonstrate the importance of local-patch masked modeling and global audio matching in extracting compact semantics from video-audio-text triplets.

In Appendix~\ref{sec: exp_more}, we also conducted extensive experiments to explore the joint encoder/decoder, and compare the video-language embedding space learned by Global Audio Matching and Video-Text Matching, separately.
These results demonstrate the effectiveness of incorporating synchronized audio into video-language pre-training in capturing more discriminative representations for both text-video and text-audio retrieval.
Furthermore, the representations extracted by the proposed GAM in our VLSA are inter-modality compact for matching pairs, while inter-modality separable and intra-modality compact for non-matching pairs.

\section{Conclusion}

In this work, we present VLSA, a novel and enhanced framework for video-language pre-training with synchronized audio that can jointly learn compact representations for video-audio-text triplets in a unified self-supervised transformer.
We introduce local-patch masked modeling to learn modality-aware local features from a joint transformer encoder.
Then, we leverage the joint encoder with global-token alignment to capture discriminative global features.
Empirical experiments on five comprehensive cross-modal retrieval benchmarks demonstrate the significant advantage of our VLSA against previous video-language pre-training approaches.
Our simple model pre-trained on only 0.9M data achieves competitive results on retrieval across text, video, and audio.
Furthermore, qualitative visualizations vividly showcase the advantage of our VLSA in learning compact visual-textual representations.

\noindent\textbf{Broader Impact.}
The proposed approach pre-trains representations of video-audio-text triplets from manually collected video datasets, which might cause the model to learn internal biases in the data. 
For instance, the model could fail to learn the correspondence between rare and noisy sounds. 
Therefore, these issues should be addressed for the deployment of real applications.

\bibliography{reference}
\bibliographystyle{unsrt}

\newpage
\appendix

\appendix
\section*{Appendix}

In this appendix, we provide significant differences between our VLSA and current video-text-audio pre-training baselines.
In addition, we compare the proposed VLSA with those baselines in terms of text-to-video retrieval on fine-tuned and zero-shot settings.
Finally, we report the quantitative comparison of computation costs with state-of-the-art methods.

\section{Video-Text-Audio Pre-training Baselines}

We conduct a comprehensive experimental study of existing approaches with video-text-audio pre-training.
Namely, we considered:

\begin{itemize}
    \item JSFusion~\cite{Yu2018AJS} (2018'ECCV): a very early sequence fusion model with multi-modal matching;
    \item MEE~\cite{Miech2018LearningAT} (2018'arXiv): a baseline with mixed embedding experts for visual-textual modalities;
    \item HT~\cite{Miech2019howto100m} (2019'ICCV): a simple baseline with two uni-modal encoders by learning text-video joint embedding;
    \item CE~\cite{Liu2019a} (2019'BMVC): a simple baseline based on a collaborative model with multiple pre-trained experts on each modality to learn a joint video-text embedding.
    \item MMV~\cite{alayrac2020self} (2020'NeurIPS): a multimodal versatile network with a deflation process to integrate text with audio-visual representations into a common embedding space.
    \item AVLnet~\cite{rouditchenko2020avlnet} (2021'Interspeech): a self-supervised baseline that learns a shared audio-visual-textual embedding space directly from raw video and text signals;
    \item ActBERT~\cite{Zhu2020ActBERTLG} (2020'CVPR): a tangled transformer with the global-local correlation between actions and object regions;
    \item HERO~\cite{li2020hero} (2020'EMNLP): a hierarchical architecture combined with cross-modal and temporal transformer;
    \item VidTranslate~\cite{Korbar2020VideoUA} (2020'arXiv): a generative method with a translation objective between modalities;
    \item UniVL~\cite{Luo2020UniVL} (2020'arXiv): a unified pre-training model for video-language understanding and generation;
    \item MMT~\cite{gabeur2020mmt} (2020'ECCV): a multi-modal transformer to aggregate per-frame visual features with temporal information;
    \item NoiseEstimation~\cite{Amrani2021NoiseEU} (2021'AAAI): a multi-modal density estimation method for learning the cross-modal correspondence;
    \item ClipBERT~\cite{lei2021less} (2021'CVPR): a generic framework based on BERT with sparsely sampled short video clips for pre-training;
    \item Frozen~\cite{Bain2021frozen} (2021'ICCV): a curriculum learning-based joint transformer with attention modeling on both space and time;
    \item SupportSet~\cite{patrick2021supportset} (2021'ICLR): a generative method by recovering caption with a weighted combination of support visual features;
    \item VATT~\cite{akbari2021vatt} (2021'NeurIPS): a unified transformer with the multi-modal contrastive loss on global features of each modality for learning the alignment of video-audio-text triplets.
    \item OA-Trans~\cite{wang2022oatrans} (2022'CVPR): an object-aware pre-training transformer with bounding boxes and tags as guidance;
    \item TVLT~\cite{tang2022tvlt} (2022'NeurIPS): a very recent visual-audio pre-training framework with masked audio/video autoencoding and contrastive modeling to align video and audio.
    \item Everything-At-Once~\cite{shvetsova2022everything} (2022'CVPR): a modality agnostic fusion transformer that integrates embeddings from three separate encoders into a fused representation in a joined multi-modal embedding space;
    \item AllinOne~\cite{wang2022allinone} (2023'CVPR): a unified transformer that learns joint representations from raw video and textual inputs.
\end{itemize}

\section{Differences between VLSA and Video-Text-Audio Pre-training Baselines}

When compared to previous video-text-audio pre-training baselines, there are three significant distinct characteristics of our VLSA for addressing video-language pre-training problems, which are highlighted as follows:

1) \noindent\textbf{Achieving competitive performance with only 0.9M triplets for pre-training.}
The major difference is that we use only 0.9M video-text-audio triplets for pre-training to learn disentangled video-text representations for retrieval. 
However, current video-text-audio pre-training approaches used 8$\sim$138M triplets, and most methods utilized all data in the whole HowTo100M~\cite{Miech2019howto100m} benchmark.
Meanwhile, some strong baselines, such as VATT~\cite{akbari2021vatt} and MMV~\cite{alayrac2020self}, combined AudioSet~\cite{gemmeke2017audioset} with HowTo100M~\cite{Miech2019howto100m} to pre-train a total of 138M data, as shown in Table~\ref{tab: exp_msrvtt_vta}.

2) \noindent\textbf{Replacing Video-Text Matching with Global Audio Matching.}
We introduce global audio matching to replace video-text matching for extracting disentangled visual-textual representations from video-text-audio triplets.
However, all previous works mainly utilized video-text matching on video frames and high-level textual semantics.
Since most frames in a video look similar in high-level textual semantics, they would bring false alignment pairs during pre-training.
Different from them, we leverage the global audio-video and audio-text matching to explicitly learn the cross-modal alignment between synchronized audio and video frames/captions, as one audio spectrogram is distinct in each audio-video/text pair.

\begin{table*}[t]
    \renewcommand\tabcolsep{8.0pt}
    \renewcommand{\arraystretch}{1.1}
	\centering
        \caption{{\bf Quantitative results of text-to-video retrieval on MSR-VTT dataset.} 
    ``Single'', ``Dual'', and ``Triple'' refer to one joint encoder, two separate, and three separate encoders.
    Bold denotes the best results. }
	\label{tab: exp_msrvtt_vta}
	\scalebox{0.64}{
	\begin{tabular}{lcccccccc}
			\toprule
			\bf Method & \bf Modalities & \bf Architecture & \bf Pre-train Datasets & \bf Data Size ($\downarrow$) & \bf R@1 ($\uparrow$) & \bf R@5 ($\uparrow$) & \bf R@10 ($\uparrow$) \\
			\midrule
            CE~\cite{Liu2019a} &  Video, Text, Audio & Triple & YouTube-8M & 8M & 20.9 & 48.8 & 62.4 \\ 
            AVLnet~\cite{rouditchenko2020avlnet} & Video, Text, Audio & Triple &  HowTo100M & 136M & \textbf{27.1} & 55.6 & 66.6 \\
            Everything-At-Once~\cite{shvetsova2022everything} & Video, Text, Audio & Triple &  HowTo100M & 136M & 23.7 & 52.1 & 63.7 \\
            \rowcolor{gray!20}
			VLSA (ours) & Video, Text, Audio & Single & AudioSet$\cap$HowTo100M & \textbf{0.9M} & \textbf{27.1} & \textbf{57.3} & \textbf{68.9} \\ \hline
			\textit{zero-shot:} \\
                VATT~\cite{akbari2021vatt} & Video, Text, Audio & Single &  AudioSet+HowTo100M & 138M & -- & -- & 29.7 \\
                MMV~\cite{alayrac2020self} & Video, Text, Audio & Triple &  AudioSet+HowTo100M & 138M & 9.3 & 23.0 & 31.1  \\
                Everything-At-Once~\cite{shvetsova2022everything} & Video, Text, Audio & Triple &  HowTo100M & 136M & 9.9 & 24.0 & 32.6 \\
                \rowcolor{gray!20}
			VLSA (ours) & Video, Text, Audio & Single & AudioSet$\cap$HowTo100M & \textbf{0.9M} & \textbf{20.4} & \textbf{40.6} & \textbf{53.8} \\
			\bottomrule
			\end{tabular}}
        \vspace{-0.5em}
\end{table*}

3) \noindent\textbf{Incorporating Local-Patch Masked Modeling for all modalities.}
We apply local-patch masked modeling on each modality through a single unified encoder and three separate decoders with shared parameters, but those video-text-audio pre-training baselines used the contrastive loss to learn the global alignment across various modalities in the joint embedding space.
They do not involve the explicit masked modeling mechanism to learn local modality-aware features across three different modalities.
In addition, they extracted local modality-aware embeddings through three separate encoders without shared parameters, except that VATT~\cite{akbari2021vatt} introduced the unified encoder to extract modality-agnostic embeddings from raw signals.
However, VATT~\cite{akbari2021vatt}  performs worse on the zero-shot text-to-video retrieval than our VLSA with local-patch masked modeling.

\section{Comparison Results with Previous Video-Text-Audio Pre-training Baselines}

Table~\ref{tab: exp_msrvtt_vta} reports the comparison results with video-text-audio pre-training baselines on fine-tuned and zero-shot text-to-video retrieval on the MSR-VTT dataset.
We can observe that the proposed VLSA achieves the best performance in terms of R@1, R@5, and R@10 compared to all baselines fine-tuned on MSR-VTT.
In particular, our VLSA significantly outperforms Everything-At-Once~\cite{shvetsova2022everything}, the recent video-text-audio pre-training approach with global alignment across three different modalities, where the performance gains of 3.4 R@1, 5.2 R@5, and 5.2 R@10 are achieved.
Meanwhile, the proposed VLSA only needs 0.9M triplets for pre-training to achieve competitive results while they used the whole HowTo100M~\cite{Miech2019howto100m} with 136M triplets.
Compared to CE~\cite{Liu2019a} pre-trained on the data with the comparable size, we achieve performance gains of 6.2 R@1, 8.5 R@5, and 6.5 R@10, as they involved neither global audio matching nor local-patch masked modeling.
These improvements validate the superiority of the proposed VLSA in enhancing video-language representations for text-video retrieval by pre-training with synchronized audio.

In addition, significant gains in zero-shot text-audio retrieval can be observed.
Compared to Everything-At-Once~\cite{shvetsova2022everything}, we achieve significant zero-shot performance gains of 10.5 R@1, 16.6 R@5, and 21.2 R@10, which implies the importance of introducing local-patch masked modeling to learn discriminative modality-aware representations across different modalities.
Meanwhile, the proposed VLSA with a unified transformer pre-trained on 0.9M video-audio-text triplets significantly outperforms MMV~\cite{alayrac2020self} pre-trained on 136M triplets and 2M video-audio pairs.
These results further demonstrate the effectiveness of our method with global audio matching in learning compact and disentangled cross-modal embeddings for zero-shot text-video retrieval.

\begin{table}[t]
\centering
\caption{{\bf Quantitative results of computation costs.} Lower is better.}
\label{tab: exp_cost}
\scalebox{0.85}{
	    \begin{tabular}{lccccccc}
			\toprule
                \multirow{2}{*}{\bf Method} & \multirow{2}{*}{\bf Params} & \bf GPU & \bf Infer \\
			 & & \bf Hours & \bf Latency (ms) \\
			\midrule
   AVLnet~\cite{rouditchenko2020avlnet} + text & 353M & -- & 2316 \\
   TVLT~\cite{tang2022tvlt} + text & 283M & -- & 2135 \\
   Frozen~\cite{Bain2021frozen} & 232M & 10580 & 1989
\\
   OA-Trans~\cite{wang2022oatrans} & 232M &  7680 & 1946\\
   \rowcolor{gray!20}
                VLSA (ours) & \textbf{156M} & \textbf{1350} & \textbf{1738} \\
			\bottomrule
	    \end{tabular}}
\end{table}

\section{Computation Costs}
For computation costs, we used 8 V100-32GB GPUs for pre-training and fine-tuning.
It should be noted that the proposed LPMM \& GAM modules in our VLSA take single pass cost during pre-training such that we bring much fewer computation costs with only 0.9M video-audio-transcript triplets.
We report quantitative comparison results of parameters, pre-training GPU hours, and inference latency in Table~\ref{tab: exp_cost}.
As can be seen, we achieve the lowest parameters compared to previous video-language pre-training baselines.
In particular, the proposed VLSA significantly decreases the parameters of OA-Trans~\cite{wang2022oatrans}, the current state-of-the-art method, by 76M.
Moreover, we achieve superior performance gains compared to TVLT~\cite{tang2022tvlt}, the current state-of-the-art video-audio pre-training baseline, which implies the importance of a joint transformer encoder in aggregating local and global visual-textual representations.

Meanwhile, our VLSA outperforms strong video-language pre-training approaches,  Frozen~\cite{Bain2021frozen} and OA-Trans~\cite{wang2022oatrans}, by large margins, where we achieve the pre-training cost reduction of 9230 GPU hours and 6330 GPU hours. 
Furthermore, when evaluating the inference latency, the proposed approach still outperforms OA-Trans~\cite{wang2022oatrans} by 208 ms. 
We also achieve highly better results against TVLT~\cite{tang2022tvlt}, the joint video-audio pre-training baseline. 
These significant cost reductions demonstrate the efficiency of our method in pre-training a joint transformer encoder on only 0.9M video-audio-transcript triplets.

\section{More Experimental Analysis}\label{sec: exp_more}

In this section, we also conducted extensive experiments to explore the joint encoder/decoder, and compare the video-language embedding space learned by Global Audio Matching and Video-Text Matching, separately.

\begin{wrapfigure}{r}{0.5\textwidth}
\centering
\vspace{-0.5em}
\begin{subfigure}{0.49\linewidth}
    \centering
    \includegraphics[width=\linewidth]{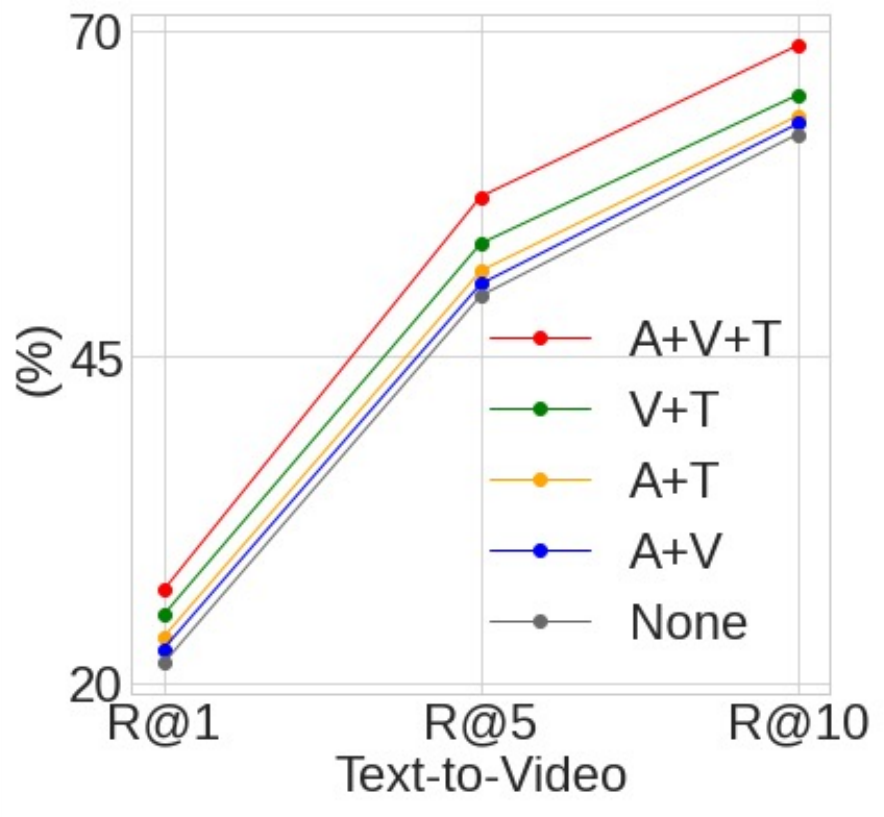}
\end{subfigure}%
\begin{subfigure}{0.49\linewidth}
    \centering
    \includegraphics[width=\linewidth]{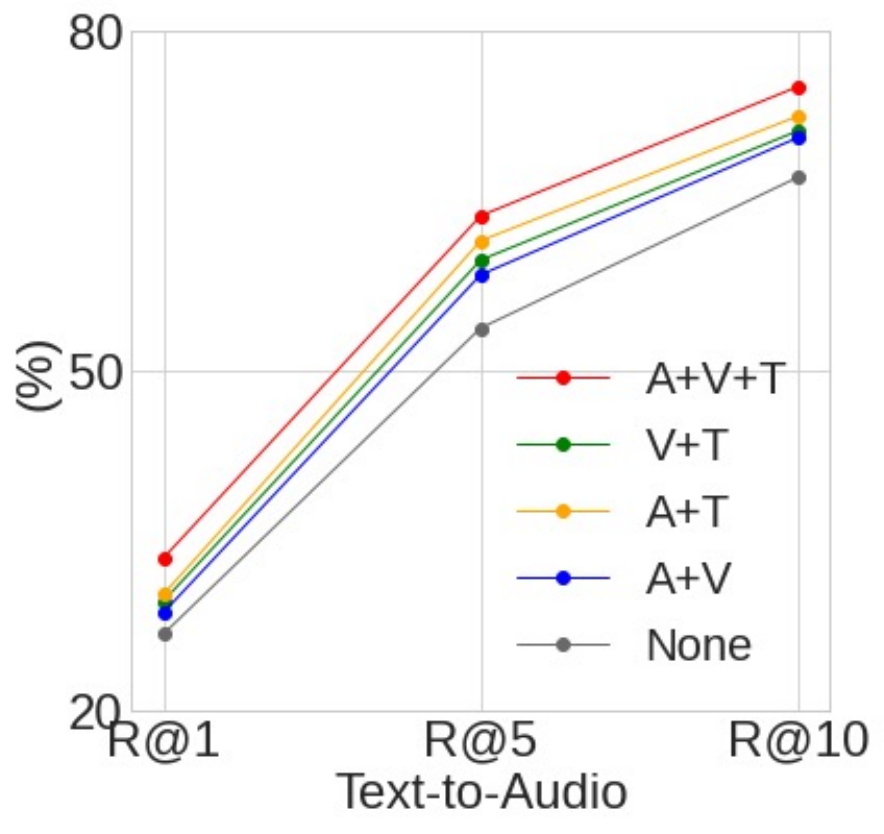}
\end{subfigure}
\vspace{-0.5em}
\caption{{\bf Effect of modality types in a single joint encoder.} A, V, and T denote audio, video, and text, respectively.}
\vspace{-0.5em}
\label{fig: ab_encoder}
\end{wrapfigure}

\noindent\textbf{Joint Encoder.}
In order to explore the effect of the modality types in a single joint encoder on both text-video and text-audio performance, we vary the combination types from $\{$A+V+T, V+T, A+T, A+V, None$\}$, where ``None'' denotes that three modality-specific encoders are used.
A, V, and T denote audio, video, and text, respectively.
The comparison results of the retrieval performance are shown in Figure~\ref{fig: ab_encoder}.
As can be seen, combining any two modalities among audio, video, and text in a joint encoder increases the results of the vanilla baseline with modality-specific encoders, which implies the importance of the proposed joint encoder in learning cross-modal representations for retrieval tasks.
In addition, adding audio to the V+T joint encoder significantly outperforms the baseline by 5.6 R@1, 7.6 R@5, 6.8R@10 for text-video, and 6.7 R@1, 9.9 R@5, 8.0R@10 for text-audio.
These results demonstrate the effectiveness of incorporating synchronized audio into video-language pre-training in capturing more discriminative representations for both text-video and text-audio retrieval.

\begin{wrapfigure}{l}{0.5\textwidth}
\centering
\begin{subfigure}{0.49\linewidth}
    \centering
    \includegraphics[width=\linewidth]{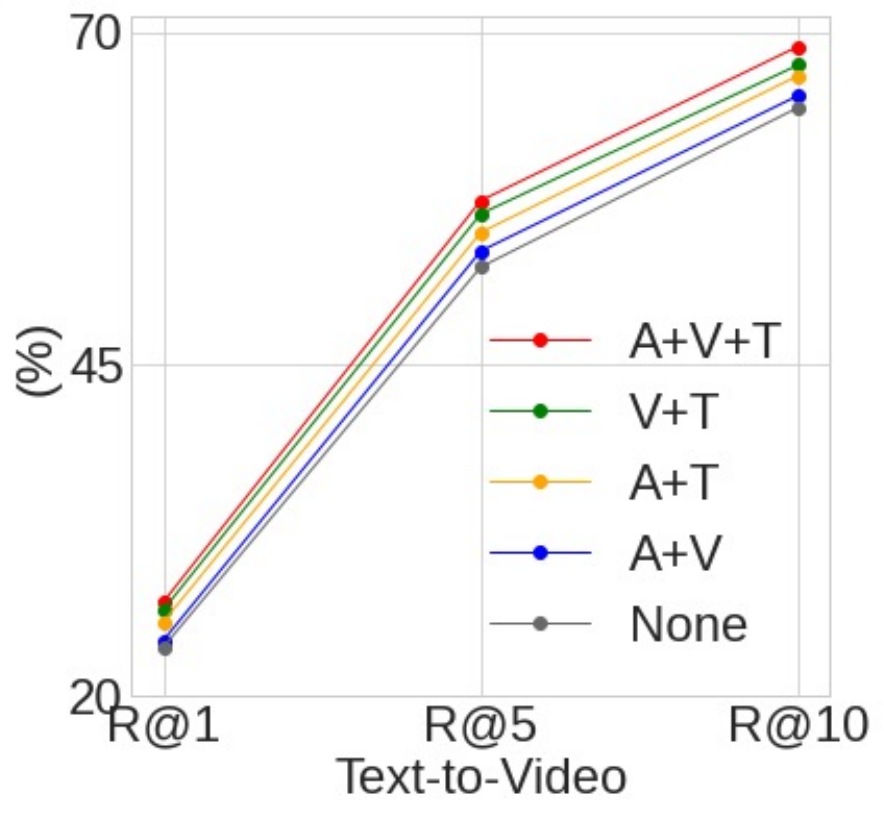}
\end{subfigure}%
\begin{subfigure}{0.49\linewidth}
    \centering
    \includegraphics[width=\linewidth]{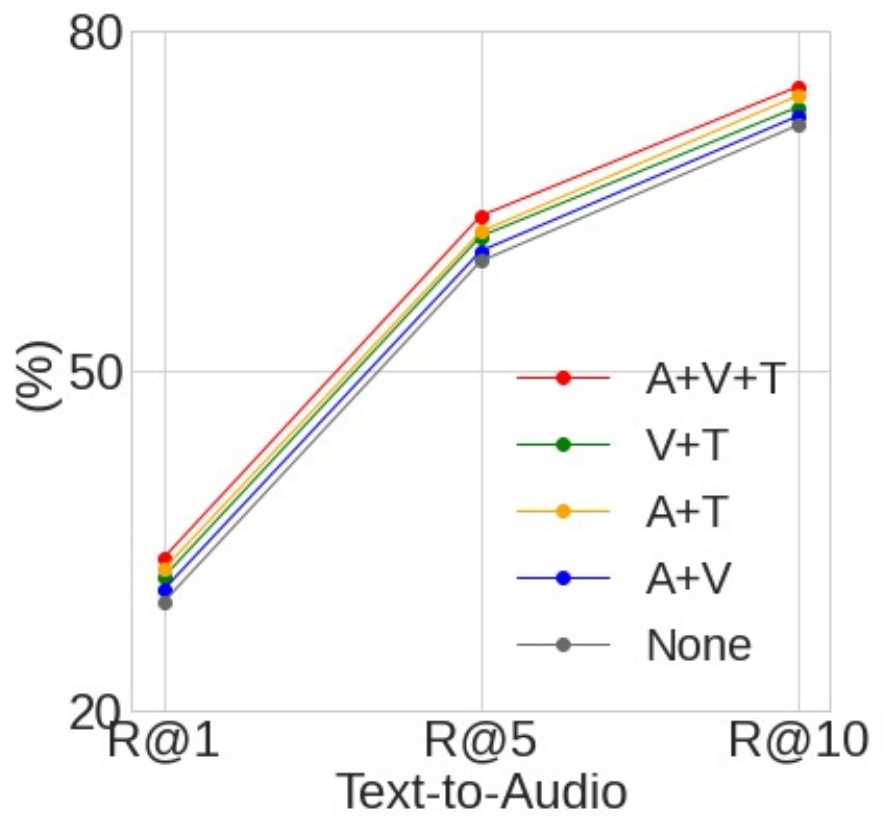}
\end{subfigure}
\vspace{-0.5em}
\caption{{\bf Effect of modality types with parameter-shared decoder.} A, V, and T denote audio, video, and text, respectively.}
\vspace{-0.5em}
\label{fig: ab_decoder}
\end{wrapfigure}

\noindent\textbf{Parameter-Shared Decoders.}
In order to explore how the parameter-shared decoder affects the final retrieval performance, we ablate the modality types from $\{$A+V+T, V+T, A+T, A+V, None$\}$, where ``None'' denotes that three separate decoders do not share parameters.
Figure~\ref{fig: ab_decoder} reports the comparison results on both text-video and text-audio retrieval.
We can observe that pre-training with parameter-shared decoders of any two combined modalities outperforms three separate decoders without parameters shared. 
In the meanwhile, using three separate decoders with parameters simultaneously shared among three modalities achieves the best performance.
This further validates the rationality of local-patch masked modeling with three separate parameter-shared decoders in enhancing video-language pre-training for retrieval tasks.

\begin{figure}
\centering
\includegraphics[width=0.75\linewidth]{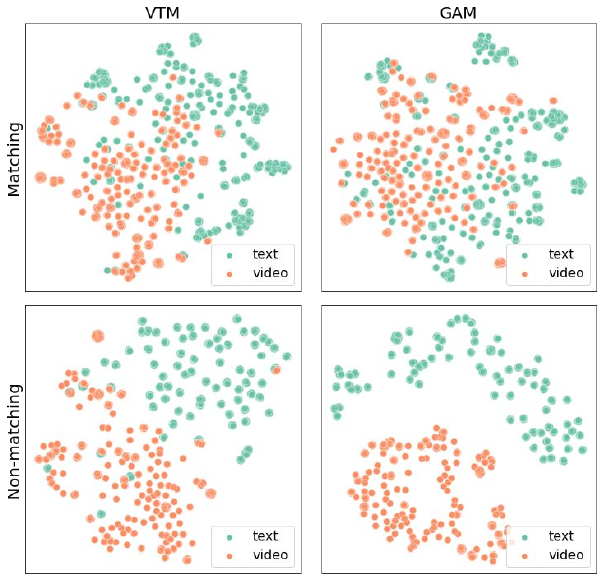}
\vspace{-0.5em}
\caption{{\bf Qualitative comparisons of visual-textual representations learned by VTM and GAM for matching (Top Row) and non-matching pairs (Bottom Row).} 
Note that each spot denotes the visual/textual feature of one video/caption, and each color refers to one modality (yellow for video, green for text).
The VLSA representations are much better.
}
\label{fig: vis_embedding}
\vspace{-0.5em}
\end{figure}

\noindent\textbf{Global Audio Matching vs. Video-Text Matching.}
Learning discriminative video-language semantic representations is essential for us to achieve higher performance in retrieval tasks. 
To better evaluate the quality of video-language representations learned by global audio matching (GAM) and video-text matching (VTM), we visualize learned visual-textual representations of 1000 randomly selected matching and non-matching pairs from MSR-VTT in a common space by t-SNE~\cite{laurens2008visualizing}, as shown in Figure~\ref{fig: vis_embedding}.
As can be seen in the last column, features extracted by the proposed GAM are inter-modality compact for matching pairs.
More importantly, GAM representations are inter-modality separable and intra-modality compact for non-matching pairs.
These meaningful visualization results further demonstrate that our VLSA successfully extracts compact visual-textual representations during pre-training. 
Note that VTM achieved 23.5 R@1, 50.6 R@5, and 61.8 R@10, which is much lower than GAM (27.1 R@1, 57.3 R@5, 68.9 R@10).

\section{Limitation}

Although the proposed VLSA achieves superior results on both text-video and text-audio retrieval, the fine-tuning gains of our approach over R@1 and R@5 on text-video retrieval are not significant. 
One possible solution is to leverage CLIP~\cite{radford2021learning} pre-trained weights and fine-grained visual features for masking (such as object bounding boxes and tags), similar to OA-Trans~\cite{wang2022oatrans} for boosting performance.
Meanwhile, we notice that if we continue training for more steps, it would be hard to see significant gains on downstream tasks. 
The primary cause is that we have a limited amount of video-audio-text triplets for training. 
Therefore, the future work is potentially to gather more triplets or to explore continual learning when it comes to new data.

\end{document}